\documentclass[letterpaper]{article} 
\usepackage{aaai2026}  
\usepackage{times}  
\usepackage{helvet}  
\usepackage{courier}  
\usepackage[hyphens]{url}  
\usepackage{graphicx} 
\usepackage{natbib}  
\usepackage{caption} 
\usepackage{algorithm}
\usepackage{algorithmic}
\usepackage{microtype}      
\usepackage{xcolor}   
\usepackage{xspace}
\newcommand{\model}[0]{DiffIER\xspace}
\usepackage{graphicx}
\usepackage{amsmath}
\usepackage{amssymb}
\usepackage{booktabs}
\usepackage{multirow}
\usepackage{caption}
\DeclareMathOperator*{\argmin}{arg\,min}
\DeclareMathOperator*{\argmax}{arg\,max}

\usepackage{newfloat}
\usepackage{listings}
\DeclareCaptionStyle{ruled}{labelfont=normalfont,labelsep=colon,strut=off} 
\floatstyle{ruled}
\newfloat{listing}{tb}{lst}{}
\floatname{listing}{Listing}
\title{DiffIER: Optimizing Diffusion Models with Iterative Error Reduction}
\author {
    Ao Chen\textsuperscript{\rm 1,\rm 2},
    Lihe Ding\textsuperscript{\rm 2},
    Tianfan Xue\textsuperscript{\rm 2}
}
\affiliations {
    \textsuperscript{\rm 1}Shanghai Jiao Tong University\\
    \textsuperscript{\rm 2} The Chinese University of Hong Kong\\
    llykevin@sjtu.edu.cn
}

\usepackage{bibentry}

\begin{document}

\twocolumn[{%
\renewcommand\twocolumn[1][]{#1}%
\maketitle
\centering
    \includegraphics[width=\textwidth]{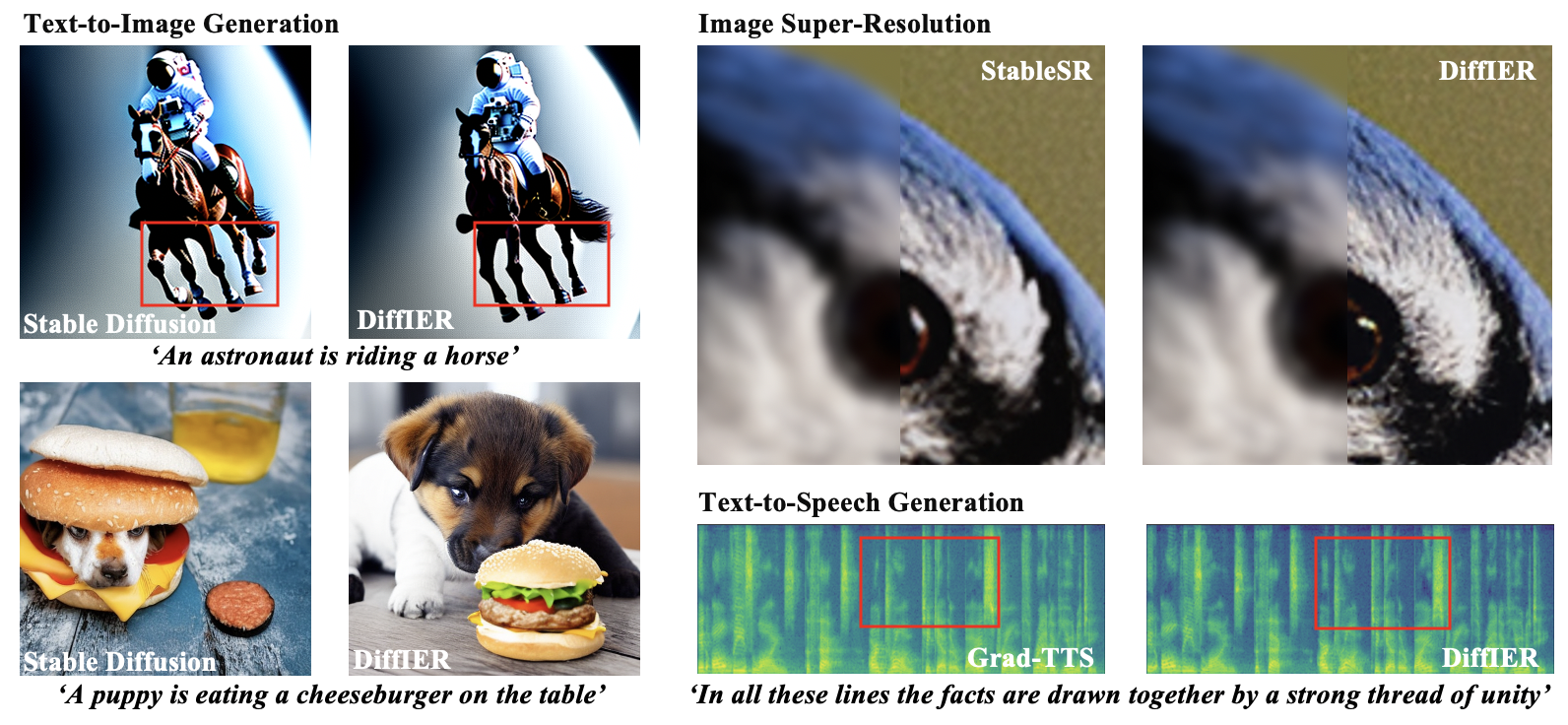}
    \captionof{figure}{We present \model, a general, training-free optimization method that achieves high-fidelity generation in diffusion models. The plug-and-play property integrates with existing diffusion-based pipelines and delivers superior performance across multiple tasks, including text-to-image generation, image super-resolution, and text-to-speech generation.}
    \label{fig:teaser}
    \vspace{1.em}

}]

\begin{abstract}
Diffusion models have demonstrated remarkable capabilities in generating high-quality samples and enhancing performance across diverse domains through Classifier-Free Guidance (CFG). However, the quality of generated samples is highly sensitive to the selection of the guidance weight. 
In this work, we identify a critical ``training-inference gap'' and we argue that it is the presence of this gap that undermines the performance of conditional generation and renders outputs highly sensitive to the guidance weight. We quantify this gap by measuring the accumulated error during the inference stage and establish a correlation between the selection of guidance weight and minimizing this gap.
Furthermore, to mitigate this gap, we propose \model, an optimization-based method for high-quality generation. We demonstrate that the accumulated error can be effectively reduced by an iterative error minimization at each step during inference. By introducing this novel plug-and-play optimization framework, we enable the optimization of errors at every single inference step and enhance generation quality.
Empirical results demonstrate that our proposed method outperforms baseline approaches in conditional generation tasks. Furthermore, the method achieves consistent success in text-to-image generation, image super-resolution, and text-to-speech generation, underscoring its versatility and potential for broad applications in future research.
\end{abstract}

\section{Introduction}
\label{introduction}

Diffusion models have demonstrated superior performance in generation across diverse domains, including image and video generation and editing, text-to-speech generation, and 3D generation \cite{song2020score, jiang2024auto, popov2021grad, rombach2022high, ding2023text}.
The probabilistic diffusion process~\cite{sohl2015deep} consists of a forward process that injects noise into a data distribution and a reverse process that recovers the noised distribution.
Theoretically, implementing the reverse process requires computing the log-likelihood of the data at every single time stamp during the forward process~\cite{sohldickstein2015deepunsupervisedlearningusing}.
However, direct calculation of data log-likelihoods is computationally intractable, \citet{song2020score} instead uses a neural network to predict score functions to approximate the gradient of the log-likelihood at the training stage and generate samples through an iterative sampler at the inference stage.

\begin{figure*}[!t]
\frenchspacing
   \includegraphics[width=\linewidth]{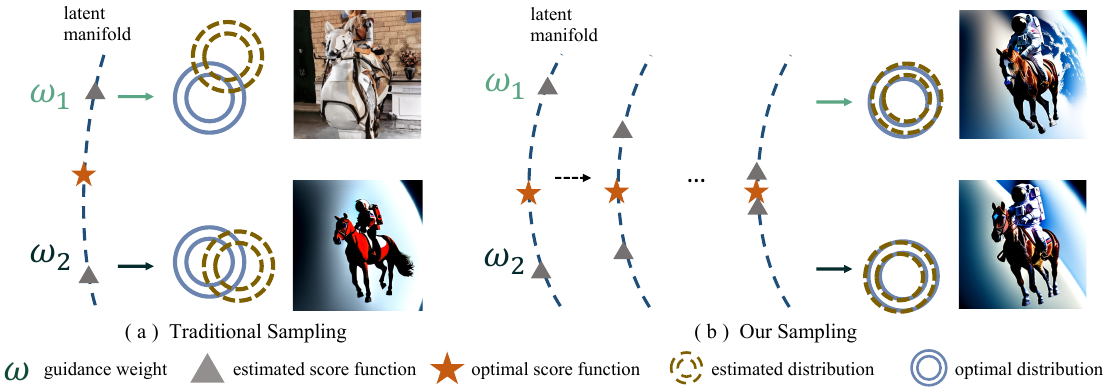}
   \caption{Illustration of how \model improves diffusion models at inference time. (a) Traditional sampling with Classifier-Free Guidance. For different input conditions, the error between the estimated score function and optimal score function can be manipulated with various guidance weights $\omega$. We use this error to measure the ``training-inference gap''. A smaller gap leads to a better performance. (b) We introduce \model, an optimization-based method to reduce the ``training-inference gap''. By ensuring the convergence of the estimated score function to the optimal value, we reduce the error, thereby improving the sampling quality and mitigating the fundamental challenges of weight selection. }
\label{fig:method}
\end{figure*}

Even with these advancements, diffusion models sometimes generate low-quality results in generation tasks, raising questions about the underlying mechanisms~\cite{dhariwal2021diffusion, sohldickstein2015deepunsupervisedlearningusing}. 
\citet{goodfellow2020generative} introduced an extra trained classifier (termed \emph{Classifier Guidance}) to boost the sampling quality. 
Then, \emph{Classifier-Free Guidance} (CFG)~\cite{ho2022classifier}, a weighted score function was proposed to improve the quality of conditional sampling while avoiding training an extra classifier. 
CFG shows a significant improvement in many tasks, including image editing~\cite{brooks2023pix2pix} and video generation~\cite{wu2023tune}. While an appropriate CFG weight can enhance sample quality, an inappropriate selection can also deteriorate it, as demonstrated in Fig.~\ref{fig:method}(a). This is because, for a specific input condition, selections of guidance weights are based on empirical comparisons and trial-and-error approaches. As a result, an excessively large weight can lead to oversaturation or artifacts in images~\cite{sadat2024eliminating}. 

The above observation raises \textit{two questions}: 1) Why does a theoretically sound score function suffers a performance drop in conditional generation tasks? 2) Why a proper selection of guidance weight is important and can improve sampling quality?
Some previous research researchers tried to explain this. For example,~\citet{bradley2024classifier} illustrated that the sampling distribution can explain the performance drop, and \citet{bradley2024classifier, wang2024analysis} tried to find a more principled way to find a better guidance weight for the second question. Still, most conclusions are empirical, and the conclusions are hard to explain all observed phenomena. Instead, in this work,  we demonstrate that the existence of a \emph{gap} between training and actual inference leads to the above two questions. 

For the first question, we found that the training-inference \emph{gap} is the primary cause of the diminished sampling quality in diffusion models.
Even diffusion models~\cite{song2020score} substitute the original gradient of log-likelihood by predicting a score function, there still exists error between the estimated score function and the optimal one, as illustrated in Fig.~\ref{fig:method}(a), resulting in a ``training-inference gap''. 
This can be quantitatively measured by the accumulated error, which is the sum of errors between the model’s predicted score function and the theoretical gradient of the log-likelihood of all steps during inference. In the experiment, we will show that by minimizing this gap, we can improve the sampling quality.

For the second question, we found that the weighted score mechanism of CFG can be viewed as a compensation for this \emph{gap} with the guidance weight. To illustrate this point, we demonstrate that there exists an optimal value for the guidance weight. A proper choice of the guidance weight can lead to a smaller derivation, which finally points to more stable and effective sampling outcomes. This observation can help address the ``roller coaster'' phenomenon observed in experiments, where the quality of image generation improves as the guidance weight increases but then declines if the weight continues to rise.

Based on these observation, we further introduce a new inference-time optimization method that effectively reduces the \emph{training-inference gap}, mitigating the fundamental challenges of CFG weight selection. We show that, although the optimal guidance weight~$\omega^*$ is  computationally inaccessible, the \textit{training-inference gap} can be effectively reduced by a new iteratively error reduction at each step. Specifically, for a given input condition, the model’s predictions may exhibit certain errors at each time step (the gap between triangle and star in Fig.~\ref{fig:method}). To reduce this error, at each time step, we use a gradient-based method to converge the score function derived from the trained model to the theoretical gradient of the log-likelihood during the inference stage. The optimized result is then utilized in the subsequent inference step iteratively, avoiding error accumulation. As illustrated in Fig.~\ref{fig:method}(b), the gap between star and triangle is reduced when iteration goes on. The advantage of this approach is that it effectively mitigates inference-phase errors under diverse input conditions. Meanwhile, if the gap itself is relatively small, the optimization cost will also be minimal, without incurring excessively heavy computational loads. By ensuring the convergence of the score function to the gradient of the log-likelihood, we minimize the approximation error, thereby significantly reducing the \emph{training-inference gap} and ultimately improving the accuracy and stability of the generation process.

Because \model is simple and training-free, it can easily adapt to any diffusion-based model. To illustrate this point, in the experiment, we have demonstrated that \model can boost the performance on three varied generation tasks. In text-to-image generation, it improves sampling quality and mitigates the variability induced by the selection of guidance weight. In image super-resolution (SR) and text-to-speech tasks, it also boosts the performance over the existing baselines, as shown in Fig.~\ref{fig:teaser} and the experimental result section. All these validate the effectiveness of the method and demonstrate its potential in many other tasks that utilize diffusion models.

\section{Related Work}

\paragraph{Generative models}
Generative models aim to synthesize new samples by learning data distributions, with prominent approaches including Generative Adversarial Networks (GANs), flow-based models, and diffusion models. 
GANs~\cite{goodfellow2020generative} generate samples via adversarial training between a generator and a discriminator, achieving breakthroughs in image synthesis. Further improvements, including Wasserstein GAN~\cite{arjovsky2017wasserstein} and StyleGAN ~\cite{karras2019style}, enhance model stability and generate quality. 
Flow-based models~\cite{kingma2018glow, dinh2016density},  leverage invertible transformations to optimize data likelihood directly, enabling exact density estimation and latent space interpolation. 
Diffusion models generate samples through a gradual denoising process, with foundational theories introduced by~\cite{sohl2015deep}. \cite{ho2020denoising} simplified this framework into denoising diffusion probabilistic models (DDPM) and \cite{song2020score} proposed a sampling scheme with score function and while \cite{song2020implicit} accelerated sampling process named denoising diffusion implicit models (DDIM). Hybrid methods like Diffusion-GAN~\cite{wang2022diffusion} combine the stability of diffusion with the efficiency of GANs. Recent advancements prioritize efficiency~\cite{liu2023instaflow, yin2024one} and multimodal capability~\cite{popov2021grad,rombach2022high,liu2023zero,poole2022dreamfusion}, driving generative models toward practical and scalable applications.

\paragraph{Guidance in Diffusion models}
Dhariwal and Nichol~\cite{dhariwal2021diffusion} introduced classifier guidance into the diffusion model and achieved superior performance in the conditional image generation task. Through training a separate classifier and replacing the estimated conditional score with a weighted alternative, this work provides an enlightening impact on the subsequent use of guidance. Classifier-free guidance (CFG)~\cite{ho2022classifier} was then introduced to prove that guidance can be performed with a pure generative model without training an extra classifier, where the guidance was expressed by a weighted sum of conditional score and unconditional score. Given the convenience of implementing CFG in practical use, generative tasks have increasingly incorporated this technology~\cite{jiang2024auto, rombach2022high, ding2023text, poole2022dreamfusion}. While the emphasis lies more on the contribution of guidance weight to downstream task performance, researchers also investigate the underlying mechanism for the differences in generative outcomes~\cite{schmidt2019generalization, ning2023input, ning2023elucidating, li2023alleviating, yuzhemanifold}. 

\paragraph{Investigation on Diffusion Guidance}
To provide an analysis on the setting of CFG weights, 
Extensive experimentation by~\cite{wang2024analysis} confirmed that monotonically increasing weight schedulers consistently improve performance. 
\cite{shenoy2024gradient}leveraged a pre-trained classifier in inference mode, dynamically determining guidance scales at each time step so that improving generation performance in both class-conditioned and text-to-image generation tasks.
\cite{karras2024guiding} raised the idea of refining a diffusion model with its inferior version. However, in practical applications, relying on the mutual optimization of the two models is of extremely high cost, and the assumptions in the reasoning process are relatively strong~\cite{dou2024diffusion}.   
\cite{chung2024cfg} identified the off-manifold phenomenon of CFG and reformulated text guidance as an inverse problem to enhance sample quality at lower guidance weights. However, these aforementioned works overlook detailed analytical exploration of CFG and do not address the underlying reasons for the changes in weight strategies across varying conditions.
To deconstruct the essence of CFG,
\cite{bradley2024classifier} proposed that CFG functions as a predictor-corrector framework and provides a certain level of analytical insight. Nevertheless, this theory does not account for the observed influence of weight adjustments on image content, nor does it explain the decline in generation quality associated with higher weights.

\section{Preliminaries}

\paragraph{Diffusion Probabilistic Model} 
\label{p3.1}
Diffusion probabilistic models firstly construct a forward process   $ q(\mathbf{x}_{1:T} | \mathbf{x}_0)$  that injects noise to a data distribution $q(\mathbf{x}_0)$, and then reverse the forward process to recover it. Given a forward noise schedule $\beta_t \in (0, 1), n = 1, \dots, T $, we define a Markov forward process~\cite{ho2020denoising}:
\begin{align}
\begin{cases}
    q(\mathbf{x}_{1:T} | \mathbf{x}_0) = \prod_{t=1}^T q(\mathbf{x}_t | \mathbf{x}_{t-1}),\\
    q(\mathbf{x}_t | \mathbf{x}_{t-1}) = \mathcal{N}(\mathbf{x}_t | \sqrt{\alpha_t} \mathbf{x}_{t-1}, \beta_t \mathbf{I}),    
\end{cases}
\label{forward process}
\end{align}
where $\mathbf{I}$ is the identity matrix, $\{\alpha_t\}$ and $\{\beta_t\}$ are scalars, and  $\alpha_t := 1 - \beta_t$. In the rest of the paper,
we adopt $q(\cdot)$ denotes the probability of both the forward diffusion process and its theoretical reverse process, conditioned on the pre-specified parameters $\{\beta_t\}$ and $p_{\theta}(\cdot)$ denotes the probability of the reverse process that is realized by a trained neural network with parameter $\theta$.

The reverse process for Eq.~\ref{forward process} is defined as a Markov process that approximates the data distribution $q(\mathbf{x}_0)$ by gradually denoising from the standard Gaussian distribution $\mathbf{x}_T \sim \mathcal{N}(0, \mathbf{I})$ through $ q(\mathbf{x}_{t-1}|\mathbf{x}_t) = \mathcal{N}(\mathbf{x}_{t-1}|\mu_t(\mathbf{x}_t), \sigma_t^2 \mathbf{I})$:
where $\mu_t(\mathbf{x}_t)$ is generally parameterized by a time-dependent score-based model $s_{\theta}(\mathbf{x}_t)$ \cite{song2020score, song2020implicit}:
\begin{equation}
\label{mu}
\begin{cases}
    \mu_t(\mathbf{x}_t) = \tilde{\mu}_t\left(\mathbf{x}_t, \frac{1}{\sqrt{\bar{\alpha}_t}}(\mathbf{x}_t + \bar{\beta}_t s_{\theta}(\mathbf{x}_t))\right),\\
    \tilde{\mu}_t(\mathbf{x}_t, \mathbf{x}_0) = \sqrt{\bar{\alpha}_{t-1}} \mathbf{x}_0 + \sqrt{\bar{\beta}_{t-1}} \cdot (\mathbf{x}_t - \sqrt{\bar{\alpha}_t} \mathbf{x}_0)/\sqrt{\bar{\beta}_t},
\end{cases}
\end{equation}
where $\bar{\alpha}_t := \Pi^t_{i=1} \alpha_i$ and $\bar{\beta}_t := 1-\bar{\alpha}_t$. 

The core of implementing the reverse process lies in computing the gradient of the log-likelihood of the data $\nabla_{\mathbf{x}_t}\log q(\mathbf{x}_t)$ at different time stamps during the forward process.
However, calculating the log-likelihood of the data at different time stamps is computationally infeasible. Instead, the key idea from ~\citet{song2020score} is to predict a score function using a neural network$s_{\theta}(\mathbf{x}_t)$,  where this score function approximates the gradient of the log-likelihood of every time stamp.
Following~\cite{song2020score}, a neural network with parameter $\theta$ is trained to approximate $s_{\theta}(\mathbf{x}_t)\approx\nabla_{\mathbf{x}_t}\log q(\mathbf{x}_t)$:
\begin{equation}
\label{loss}
\theta^* = \argmin_{\theta} \mathbf{E}_{\mathbf{x}_t\sim q(\mathbf{x}_t)}\left[||s_{\theta}(\mathbf{x}_t)-\nabla_{\mathbf{x}_t}\log q(\mathbf{x}_t)||^2\right].
\end{equation}
Once the optimal score function~$\theta^*$ is achieved, the reverse process can be solved by replacing the gradient of data log-likelihood ~$\nabla_{\mathbf{x}_t}\log q(\mathbf{x}_t)$ with a predicted score function~$s^*_{\theta}(\mathbf{x}_t)$, achieving the goal for sampling from distribution $\mathbf{x}_T \sim \mathcal{N}(0, \mathbf{I})$~\cite{song2020score}.


\paragraph{Classifier-Free Guidance.} The diffusion model above is for unconditional generation, while conditional generation is more useful in practice. To achieve this, we can incorporate a condition $c$ and sample from $q(\mathbf{x}_t|c)$. \citet{dhariwal2021diffusion} proposed to decompose the conditional score function using Bayes' Theorem as $\nabla_{\mathbf{x}_t}\log q(\mathbf{x}_t|c) = \nabla_{\mathbf{x}_t}\log q(\mathbf{x}_t) + \nabla_{\mathbf{x}_t}\log q(c|\mathbf{x}_t)$.
This decomposition leads to the well-known \textit{Classifier Guidance (CG)}, which introduces a weight $\omega > 0$ to emphasize the condition $c$:
\begin{equation}
\label{cg}
    s^{\text{cg}}_{\theta, \omega} (\mathbf{x}_t, c)= s_{\theta}(\mathbf{x}_t) + (\omega+1)\cdot \nabla_{\mathbf{x}_t}\log q(c|\mathbf{x}_t).
\end{equation}
However, this approach comes at the cost of training an additional classifier~$q(c|\mathbf{x}_t)$, which is often impractical. To address this limitation, \citet{ho2022classifier} further analyze the classifier by expressing it as $\nabla_{\mathbf{x}_t}\log q(c|\mathbf{x}_t) = \nabla_{\mathbf{x}_t}\log q(\mathbf{x}_t, c) - \nabla_{\mathbf{x}_t}\log q(\mathbf{x}_t)$, 
through training a diffusion network to jointly estimate both the conditional score \( \nabla_{\mathbf{x}_t}\log q(\mathbf{x}_t| c) \) and the unconditional score \( \nabla_{\mathbf{x}_t}\log q(\mathbf{x}_t| \varnothing) \), where $\varnothing$ means setting the input condition as null.
This results in the \textit{Classifier-Free Guidance (CFG)}:
\begin{equation}
\label{cfg}
    s^{\text{cfg}}_{\theta, \omega} (\mathbf{x}_t, c)= s_{\theta}(\mathbf{x}_t, \varnothing) + \omega \cdot \left(s_{\theta}(\mathbf{x}_t, c) - s_{\theta}(\mathbf{x}_t, \varnothing)\right),
\end{equation}
where $\omega$ serves as a weight (usually $\omega > 1$ and $\omega =1$ indicates no CFG) to balance the conditional and unconditional components. The choice of CFG weight $\omega$ is often ad-hoc, but it is critical for high-quality generation.

\section{Methodology}
 In this section, we first present our new perspective on CFG for a comprehensive understanding of this widely used mechanism in diffusion models. Then, we propose an optimization method to reduce the ``training-inference gap'' on current diffusion models. For brevity, we omit the derivations of some equations and refer the interested readers to the supplementary material.

\subsection{Relation between CFG and Training-Inference Gap} 
To obtain the optimal reverse process of a diffusion model, ~\citet{bao2022analytic} introduced analytic forms of mean $\mu^*_t(\mathbf{x}_t)$, as expressed in Eq.~\ref{paper_optimal_mu}:
\begin{equation}
\label{paper_optimal_mu}
    \mu_t^*(\mathbf{x}_t) = \tilde{\mu}_t\left(\mathbf{x}_t, \frac{1}{\sqrt{\bar{\alpha}_t}}\left(\mathbf{x}_t+\bar{\beta}_t\nabla_{\mathbf{x}_t}\log q(\mathbf{x}_t|c)\right)\right),
\end{equation}
where the formulation of $\tilde{\mu}$ in the equation follows the definition given in Eq.~\ref{mu} above. 

To accomplish conditional sampling in reverse process with CFG, \citet{ho2022classifier} replaces the gradient of data log-likelihood~$\nabla_{\mathbf{x}_t}\log q(\mathbf{x}_t|c)$ with a weighted predicted score function~$s^{\text{cfg}}_{\theta, \omega} (\mathbf{x}_t, c)$ by the neural network. Here, we propose a re-evaluation of the deviation between the theoretically value in the reverse process and the CFG-based one, expressed as
:
\begin{equation}
\label{mean_bias}
    \mathbf{E}_{\mathbf{x}_t}\left\|\mu_t^*(\mathbf{x}_t) - \mu_t^{\text{cfg}}(\mathbf{x}_t)\right\|^2,
\end{equation}
where $\mathbf{E}$ denotes the mathematical expectation and $\mu_t^{\text{cfg}}(\mathbf{x}_t)$ means replacing $\nabla_{\mathbf{x}_t}\log q(\mathbf{x}_t| c)$ with $s^{\text{cfg}}_{\theta, \omega} (\mathbf{x}_t, c)$  in Eq.~\ref{paper_optimal_mu} during conditional sampling cases.
Eq.~\ref{mean_bias} is reduced to the following expression $w.r.t$ the guidance weight $\omega$:
\begin{equation}
    L(\omega) \triangleq \mathbf{E}_{\mathbf{x}_t}\left\|s_{\theta, \omega}^{\text{cfg}}(\mathbf{x}_t, c) - \nabla_{\mathbf{x}_t}\log q(\mathbf{x}_t|c)\right\|^2.
\end{equation}
Then,
by strictly defining $s_{\theta}(\mathbf{x}_t, c)\approx\nabla_{\mathbf{x}_t}\log q(\mathbf{x}_t|c)$ as $\nabla_{\mathbf{x}_t}\log q(\mathbf{x}_t|c) \triangleq s_{\theta}(\mathbf{x}_t, c) + e_{t,c}$, we reach the optimal CFG weight $\omega^*$  as follows  (Details in Supplementary A.1) :
\begin{equation}
    \omega^* = 1 + \mathbf{E}_{\mathbf{x}_t}\left[\frac{\left(s_{\theta}(\mathbf{x}_t, c) - s_{\theta}(\mathbf{x}_t, \varnothing)\right)e_{t,c}}{\left(s_{\theta}(\mathbf{x}_t, c) - s_{\theta}(\mathbf{x}_t, \varnothing)\right)^2}\right],
    \label{eq:opt_w}
\end{equation}
where $e_{t,c}$ donates the error between $s_{\theta}(\mathbf{x}_t, c)$ and $\nabla_{\mathbf{x}_t}\log q(\mathbf{x}_t|c)$ under a fixed condition $c$ at time step $t$ mentioned in Sec.~\ref{introduction}.
Apparently, the optimal guidance weight~$\omega^*$ equals to 1 when the error~$e_{t,c}=0$.
The optimal CFG weight equation Eq.~\ref{eq:opt_w} shares following observations:
\begin{itemize}
\item The deviation(Eq.\ref{mean_bias}) is related to the guidance weight~$\omega$. There occurs a theoretically optimal value~$\omega^*$, $w.r.t$ time step $t$. A better choice of $\omega$ leads to a smaller deviation at every single time step, leading to a more stable and effective sampling outcome. 
\item The optimal guidance weight~$\omega^*$ equals to 1 (means no CFG) when the error~$e_{t,c}\triangleq \nabla_{\mathbf{x}_t}\log q(\mathbf{x}_t|c) - s_{\theta}(\mathbf{x}_t, c) =0$. As a result, the introduction of CFG constitutes a compensatory mechanism for the approximation error~$e_{t,c}$.
\item Considering introducing the CFG (set $\omega > 1$) can lead to a better generation result, there exists an error~$e_{t,c}\neq0$. We define the accumulating error of all sampling steps results in the ``training-inference gap''.
\item The theoretic optimal guidance weight $\omega^*$ in Eq. ~\ref{eq:opt_w} highly depends on the input condition $c$ and it is hard to calculate in practice. This explains why finding a global optimal CFG weight is impossible.
\end{itemize}


The analysis above shows 
that the occurrence of this gap leads to a drop in sampling quality. To tackle this challenge, in the next subsection, we will propose \model, an inference-time optimization method that reduces the accumulated error and improves generation quality.

\subsection{\model}
\label{Terminal Filtering}


To reduce the ``training-inference gap'' described above, we introduce an inference-time optimization method that effectively decreases the gap, mitigating the fundamental challenges of CFG. 
An ideal solution is to directly calculate the gradient of the log-likelihood $\nabla_{\mathbf{x}_t}\log q(\mathbf{x}_t| c)$ for a fixed condition $c$, which guarantees to minimize the error at any time step $t$. However, calculating the log-likelihood is computationally infeasible.

Therefore, instead of calculating the precise log likelihood to reach the optimal guidance weight $\omega^*$, we directly optimize the accumulative error itself.
We reveal that the accumulated error can be iteratively reduced through reducing the error of every single step under a fixed condition $c$ with  mathematical derivation (Details in Supplementary A.2):
\begin{equation}
    p_{\theta^*}(\mathbf{x}_{t-1}|\mathbf{x}_t) = \mathbf{E}_{p_{\theta^*}} q(\mathbf{x}_{t-1}|\mathbf{x}_{0:T-1/t-1}, \mathbf{x}_T),
\end{equation}
where $p_{\theta}(\cdot)$ means the predicted distribution from a trained network with parameters $\theta$. $t \in {1, 2, ..., T}$ and $\mathbf{x}_{0:T-1/t-1}$ represents $\{\mathbf{x}_0, \mathbf{x}_1, .., \mathbf{x}_{T-1}\}$ excluding $\mathbf{x}_{t-1}$. For an existing diffusion model such as LDM~\cite{rombach2022high}, we proposed an optimization strategy with the 
Eq.~\ref{opt_target} as the objective function:
\begin{equation}
\label{opt_target}
\hspace{-9pt} L = \left| \rule{0pt}{10pt} \right| p_{\theta}(\mathbf{x}_{t-1}|\mathbf{x}_t) - \mathbf{E}_{p_{\theta}} q(\mathbf{x}_{t-1}|\mathbf{x}_{0:T-1/t-1}, \mathbf{x}_T) \left| \rule{0pt}{10pt} \right|^2.
\end{equation}

We want to emphasize two points about this objective function. First, the time index $t$ in the reverse process starts from $T$. Considering that the target in the forward diffusion process is a Gaussian distribution, the initial step of optimization is also chosen to be Gaussian. Second, the computation of the expectation in the second term of this expression is based on the probability of the inference process. From an implementation perspective,  the optimization result at step $t$ will be iteratively used in the computation at step $t-1$.

Finally, following \cite{song2020score}, we ultimately simplify the optimization objective as (Details in Supplementary):
\begin{equation}
\label{opt_loss}
    L = \left| \rule{0pt}{10pt} \right| \epsilon^{cfg}_{\theta, \omega}(\mathbf{x}_t, c)-\epsilon \left| \rule{0pt}{10pt} \right|^2,
\end{equation}
where $\epsilon^{cfg}_{\theta, \omega}(\mathbf{x}_t, c)$ represents output from the pre-trained model at time step $t$ with input condition $c$. 
We propose the detailed algorithm of the proposed \model in Alg.~\ref{alg:optimization}, where we take the DDIM sampling method\cite{song2020implicit} as an example.

\begin{algorithm}[h]
\caption{\model: Optimizing Diffusion Models with Iterative Error Reduction}
\label{alg:optimization}
\begin{algorithmic}[1]
\STATE Input: classifier-free guidance weight: $\omega$ 
\STATE ~~~~~~~~~~~gradient scale: $\eta = 5e-2$
\STATE ~~~~~~~~~~~convergence threshold: $1e-3$
\STATE $\mathbf{x}_T \sim \mathcal{N}(0, \mathbf{I})$
\FOR{$t = T$ to $1$}
    \STATE $ \epsilon^{\text{cfg}}_{\theta, \omega} (\mathbf{x}_t, c)= \epsilon_{\theta}(\mathbf{x}_t, \varnothing) + \omega \cdot \left(\epsilon_{\theta}(\mathbf{x}_t, c) - \epsilon_{\theta}(\mathbf{x}_t, \varnothing)\right)$
    \WHILE{not convergence}
        \STATE $\epsilon \sim \mathcal{N}(0, I)$
        \STATE L = $\left \| \epsilon^{cfg}_{\theta, \omega}(\mathbf{x}_t, c)-\epsilon \right  \|^2$
        \STATE $\epsilon^{cfg}_{\theta, \omega}(\mathbf{x}_t, c) = \epsilon^{cfg}_{\theta, \omega}(\mathbf{x}_t, c)+\eta \cdot \nabla_{\epsilon^{cfg}_{\theta, \omega}(\mathbf{x}_t, c)} $L
    \ENDWHILE
    \STATE $x_{t-1}$ = DDIM Sampler$\left(\mathbf{x}_t, \epsilon^{cfg}_{\theta, \omega}(\mathbf{x}_t, t)\right)$
\ENDFOR
\STATE \textbf{return} $\mathbf{x}_0$
\end{algorithmic}
\end{algorithm}

\section{Experiments}
In this section, we first introduce the experimental setups and provide extensive experimental results to demonstrate the superiority of this work.  
\subsection{Experimental Setup}

\paragraph{Datasets and Baselines}
We evaluate our plug-and-play method on synthetic and real-world datasets across multiple tasks. 
1) For \textit{conditional sampling}, we employ LDM trained on ImageNet~\cite{deng2009imagenet}, using category labels as conditions. Additionally, we compare results with LDM~\cite{rombach2022high} using descriptive prompts for \textit{text-to-image generation}, validating our method’s effectiveness across conditioning paradigms. 
2) We explore \textit{image super-resolution (SR)} by comparing our method with StableSR~\cite{wang2024exploiting}, which leverages generative priors to overcome fixed-size limitations. We manage samples from  DIV2K~\cite{agustsson2017ntire}, LSUN-bedroom~\cite{yu2015lsun}, LSUN-church~\cite{yu2015lsun} and ImageNet~\cite{deng2009imagenet} datasets with Gaussian-blurred LR-HR pairs, rigorously evaluating resolution enhancement capabilities. 
3) Finally, to assess \textit{generalization ability}, we benchmark against Grad-TTS~\cite{popov2021grad}, a text-to-speech model that generates audio samples aligned with text input. Our evaluation uses manually designed text inputs to demonstrate robustness, versus samples generated by Luvvoice~\cite{luvvoice_com}, Text2Speech~\cite{text2speech_org}, TTS-online~\cite{texttospeech_online}, TTSMaker~\cite{ttsmaker_cn}.
All comparisons are conducted using pre-trained models as baselines.

\paragraph{Evaluation metrics}
We first conduct a quantitative comparison in conditional image generation, using established metrics: Inception Score (IS)~\cite{barratt2018note}, 
and Precision-and-Recall~\cite{kynkaanniemi2019improved}, implemented with the Torch-Fidelity library for consistency. Comparable results on the image super-resolution task are also demonstrated with CLIP-IQA~\cite{wang2023exploring}, DeQA-Score~\cite{deqa_score} and MUSIQ~\cite{ke2021musiq} to evaluate the perceptual quality of generated images. PSNR and Lpips scores are also reported for reference. Besides, we also provide results of text-to-speech with PESQ. To assess generalization, we provide visualized results for text-to-image generation, text-to-speech synthesis, and image super-resolution, offering an intuitive qualitative understanding. 


\setlength{\tabcolsep}{5pt}
\begin{table}[ht]
\caption{Quantitative comparison of conditional sampling on Imagenet dataset versus Langevin Dynamics (LD) sampler~\cite{song2020score}. We generate 5K samples on each category for comparison.}
\centering
\frenchspacing
\resizebox{0.48\textwidth}{!}{
\begin{tabular}{cccc|ccc}
\toprule
Method & Cate. & IS $\uparrow$ & Prec. $\uparrow$ & Cate. & IS $\uparrow$ & Prec. $\uparrow$ \\
\toprule
 LD & \multirow{2}{*}{101} & 1.164& 0.730 & \multirow{2}{*}{167} & 1.060 & 0.641   \\
 \model & & \textbf{1.168}& \textbf{0.748} &  &  \textbf{1.063} & \textbf{0.699}  \\
  \hline
 LD & \multirow{2}{*}{187} & 1.027 & 0.944  & \multirow{2}{*}{267} & 1.060  & 0.749 \\
 \model & & \textbf{1.168}& \textbf{0.748} & & \textbf{1.005}& \textbf{0.927}  \\
   \hline
 LD & \multirow{2}{*}{448} & 1.005& 0.923&  \multirow{2}{*}{736} & 1.012& 0.961\\
 \model & & \textbf{1.064} & \textbf{0.773}& & \textbf{1.015}& \textbf{0.966}\\
    \hline
 LD & \multirow{2}{*}{879} & 1.122& 0.908&  \multirow{2}{*}{992} & 1.016& 0.918\\
 \model & & \textbf{1.127}& \textbf{0.909}& & \textbf{1.017}& \textbf{0.928}\\
\bottomrule
\end{tabular}
}
\label{tab:conditional_sampling}
\end{table}

\subsection{Quantitative Comparisons}

\paragraph{Conditional Image Generation.} As discussed in~\cite{song2020score, bradley2024classifier}, the predictor-corrector sampling strategy can be regarded as an enhancement to conventional conditional diffusion samplers. To validate the effectiveness of our method, we provide numerical comparisons with~\cite{song2020score} in Tab.~\ref{tab:conditional_sampling}, conducted on the ImageNet~\cite{deng2009imagenet} dataset.
For efficiency and operational feasibility, we randomly select various categories to generate corresponding samples. As shown in Tab.~\ref{tab:conditional_sampling}, our approach consistently outperforms the baseline across all evaluated metrics. 


\setlength{\tabcolsep}{5pt}
\begin{table}[ht]
\caption{Quantitative comparison of image super-resolution with StableSR~\cite{wang2024exploiting}.(L:Lpips, P:PSNR, M:MUSIQ, C:Clip-IQA, D:DeQA-Score)}
\centering
\frenchspacing
\resizebox{0.48\textwidth}{!}{
\begin{tabular}{cccccc}
\toprule
Metrics & Method & DIV2K & LSUN-b & LSUN-c & ImageNet \\
\toprule
  \multirow{2}{*}{L $\downarrow$}& StableSR  & 0.265& 0.314 & 0.305 &  0.327   \\
  & \model & \textbf{0.253}& \textbf{20.69} & \textbf{0.297} &  \textbf{0.306}   \\
  \hline
  \multirow{2}{*}{P $\uparrow$} & StableSR & 20.24 & 23.83  & 20.51 & 22.88  \\
  & \model & \textbf{20.69} & \textbf{24.15} & \textbf{20.96} & \textbf{23.03}  \\
  \hline
  \multirow{2}{*}{M$\uparrow$} & StableSR & 68.13& 58.22& 52.51 & 57.86\\
  & \model & \textbf{70.46}& \textbf{20.96}& \textbf{53.55} & \textbf{59.71}\\
   \hline
 \multirow{2}{*}{C $\uparrow$} & StableSR &  0.607& 0.310& 0.387 & 0.523\\
  & \model & \textbf{0.626}& \textbf{0.321}& \textbf{0.392} & \textbf{0.604}\\
     \hline
 \multirow{2}{*}{D$\uparrow$} & StableSR & 3.032& 2.782&  2.493 & 2.796\\
  & \model & \textbf{3.141}& \textbf{2.945}& \textbf{2.574} & \textbf{2.881}\\
\bottomrule
\end{tabular}
}
\label{tab:ImageSR}
\end{table}

\paragraph{Image Super-Resolution.}
Table \ref{tab:ImageSR} illustrates the generalization capability of the proposed \model in the image super-resolution (SR) task across samples from various datasets.
As demonstrated in the table, our method achieves consistent superiority over the baseline across all evaluated metrics. 
This consistent improvement highlights the strong generalization ability of our approach, ensuring reliable and uniform performance during real-world deployment.


\begin{table}[htp]
\caption{Quantitative comparison of text-to-speech with GradTTS~\cite{popov2021grad}.}
\centering
\frenchspacing
\begin{tabular}{cccc}
\toprule
Metric & Method & LuvVoice  & Text2Speech  \\
\toprule
  \multirow{2}{*}{PESQ $\uparrow$} & GradTTS  & 1.042& 1.032  \\
  & \model & \textbf{1.044}& \textbf{1.052}   \\
\toprule
Metric & Method & TTS-online& TTSMaker  \\
\toprule
  \multirow{2}{*}{PESQ $\uparrow$} & GradTTS  & 1.030 &  1.034  \\
  & \model & \textbf{1.038} &  \textbf{1.041}   \\
\bottomrule
\end{tabular}
\label{supplytab:tts}
\end{table}

\paragraph{Text-to-speech.}

This work illustrates the generalization capability of the proposed \model in the text-to-speech (TTS) task across diverse sources. Our method consistently outperforms the baseline models in key PESQ evaluation metrics (in Tab.~\ref{supplytab:tts}). These results collectively validate the robustness and potential of our algorithm in addressing cross-modal tasks.

\subsection{Qualitative Comparisons}


\paragraph{Text-to-image Generation.} 
\begin{figure}[!ht]
\centering
\frenchspacing
   \includegraphics[width=\linewidth]{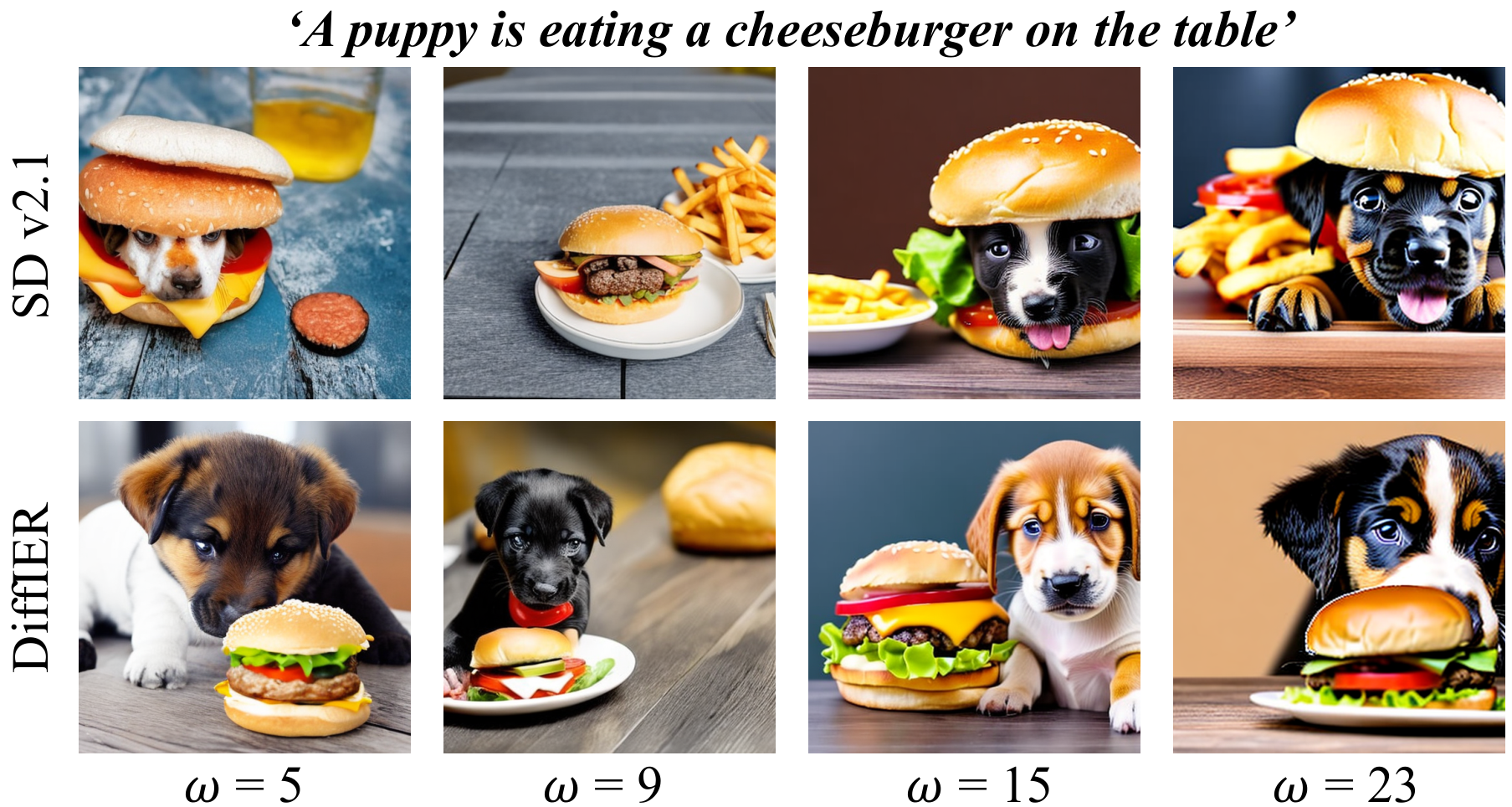}
   \caption{Results on text-to-image task, compared with Stable Diffusion (SD v2.1). For a given text prompt 'A puppy is eating a cheeseburger on the table', the generation results of SD are influenced by diverse guidance weight $\omega$ settings. Generated images from SD cannot align closely with the prompt and exhibit visual artifacts. With the application of our proposed method \model, the generated image demonstrates improved alignment with the textual prompt, and quality has been improved, mitigating the challenging weight setting of CFG mechanism.}
\label{fig:stable_diffusion}
\end{figure}

\begin{figure*}[!htp]
\centering
\frenchspacing
   \includegraphics[width=0.9\linewidth]{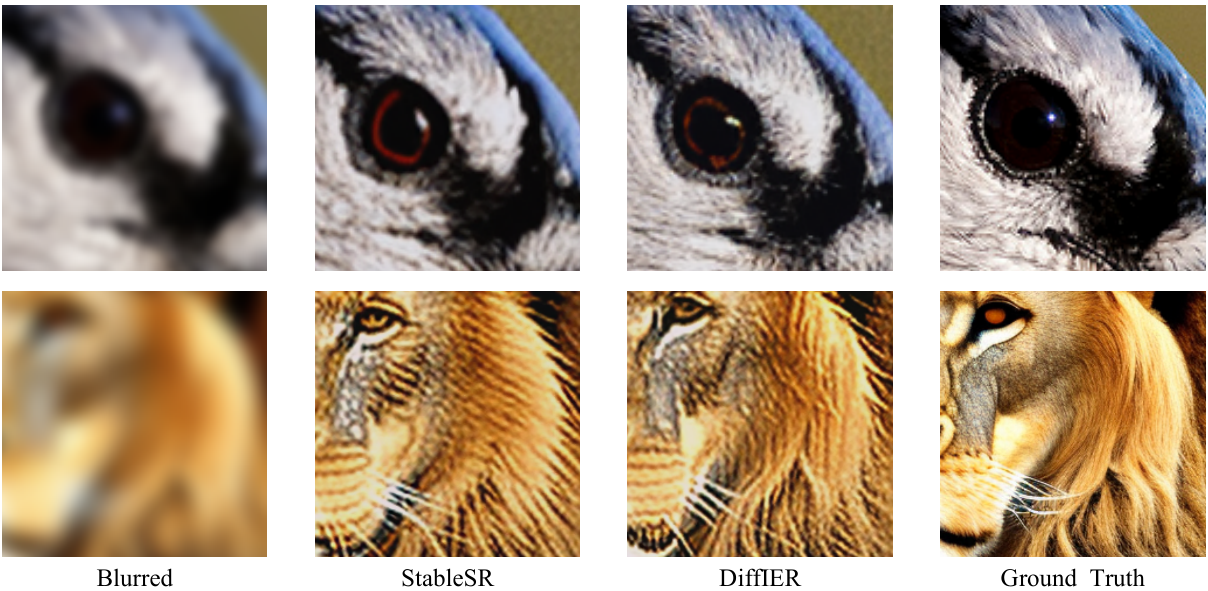}
   \caption{Results on image super-resolution task, compared with StableSR. In comparison to StableSR, our method can produce a more faithful restoration of image details.}
\label{fig:imagesr}
\end{figure*}

We evaluate the performance of our method using the pre-trained StableDiffusion ~\cite{rombach2022high} (SD v2.1) as the baseline. As illustrated in Fig.~\ref{fig:stable_diffusion},  with the prompt ``A puppy is eating a cheeseburger on the table'', while SD can generate samples, the results of diverse guidance weights cannot align with the prompt and exhibit visual artifacts. 
In contrast, our method significantly improves image quality by removing artifacts and producing more natural and coherent content. Additionally, the consistency between the generated image and the input prompt is enhanced, mitigating the fundamental challenges of CFG, as shown in Fig.~\ref{fig:stable_diffusion}. 
More visible results in Supplementary C.1.


\paragraph{Diffusion-based Image Super-Resolution.}
We provide qualitative results in Fig.~\ref{fig:imagesr}.
It can be seen that our work outperforms the baseline. When taking the images that are blurred by a Gaussian kernel as the input, our method can effectively and correctly recover the details of objects, including the feather patterns of a bird, the mane of a lion. With the ground truth images, the SR results of the baseline contain a number of artifacts, although the images become clearer. However, our method can restore the blurred part to closely align with the ground truth. This is crucial in SR tasks that effectively restore information of real-world images More visible results in Supplementary C.2.

\begin{figure}[!htp]
\centering
\frenchspacing
   \includegraphics[width=\linewidth]{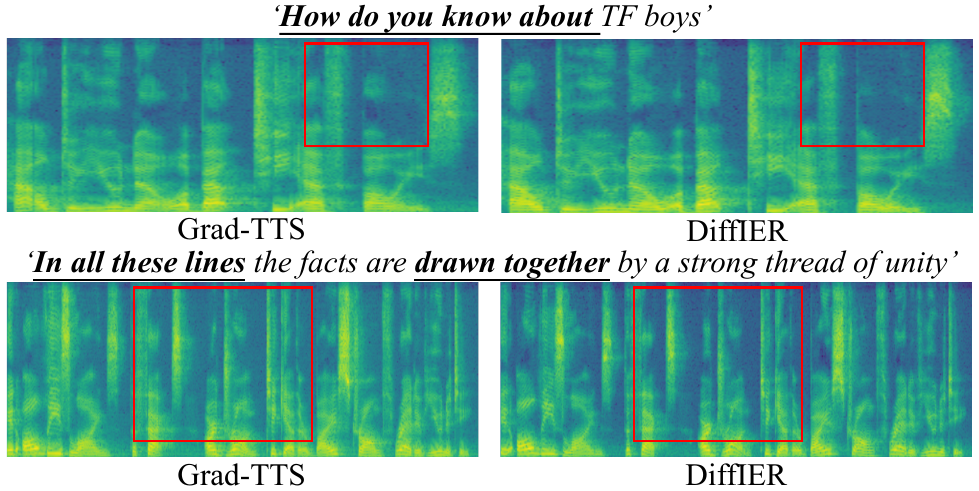}
   \caption{Results on text-to-speech task, compared with Grad-TTS. In comparison to Grad-TTS, our method achieves a higher signal-to-noise ratio in key syllables and a more natural and realistic tonal quality.} 
\label{fig:grad_tts}
\end{figure}

\paragraph{Diffusion-based Text-to-Speech Generation.} We evaluate the proposed method in Text-to-Speech (TTS) generation, using Grad-TTS~\cite{popov2021grad} as the baseline. Grad-TTS employs a score-based decoder to generate mel-spectrograms, aligning with our method’s strengths. 
As shown in Fig.~\ref{fig:grad_tts}, while both the baseline and our method can generate speech for various input text prompts, our method achieves a higher signal-to-noise ratio in key syllables. Additionally, the samples produced by our method exhibit enhanced perceptual clarity in enunciation and a more natural and realistic tonal quality.
These results validate the theoretical feasibility and generalization of our approach, demonstrating its effectiveness in cross-modal tasks. More visible results and audio samples in \textbf{Supplementary C.3}.

\section{Conclusion}
The accumulated error during the inference stage underlies the ``training-inference gap'', which is a key factor contributing to the quality deterioration in conditional sampling tasks with diffusion models. Through mathematical derivation, we demonstrate that the weight of Classifier-Free Guidance (CFG) modulates this gap during inference, and the ``training-inference gap'' can be iteratively minimized at each step. To address this issue, we propose \model, an inference-time optimization method designed to reduce the training-inference gap and mitigate the fundamental challenges associated with CFG. Our approach employs a gradient-based optimization technique to iteratively converge the error at each step, which is then utilized in subsequent inference steps. This process effectively reduces the accumulated error and significantly enhances generation quality. Empirical results highlight the effectiveness and versatility of our method, demonstrating its potential for broad applications in future research. In future work, we will explore the potential of this method to be applied to other generative architectures.


\bibliography{aaai2026}

\newpage
\appendix
\section{Technical Appendices and Supplementary Material}

\paragraph{Computational resources}

The method proposed in this work does not involve a training procedure, and the related computational experiments are conducted in the inference stage based on public models. On the hardware front, an NVIDIA 3090 GPU is employed, and for the software side, the PyTorch machine learning framework serves as the foundation.

\subsection{Detailed process of Sec. 4.1}
In this section, we present more details on the conclusion of Section 4.1. 
\citet{bao2022analytic} concluded the optimal mean value at each time step $t$ during inference stage: 
\begin{equation}
\begin{aligned}
\label{optimal_mu}
    \mu_t^*(\mathbf{x}_t) &= \tilde{\mu}_t\left(\mathbf{x}_t, \frac{1}{\sqrt{\bar{\alpha}_t}}\left(\mathbf{x}_t+\bar{\beta}_t\nabla_{x_t}\log q(\mathbf{x}_t|c)\right)\right)\\
    &= \left(\sqrt{\bar{\alpha}_{t-1}}-\frac{\sqrt{\bar{\beta}_{t-1} \bar{\alpha}_t}}{\sqrt{\bar{\beta}_t}}\right) \\
    &~~~~\cdot \left(\frac{1}{\sqrt{\bar{\alpha}_t}}x_t + \frac{\bar{\beta}_t}{\sqrt{\bar{\alpha}_t}}\nabla_{x_t}\log q(\mathbf{x}_t|c) \right) + \sqrt{\bar{\beta}_{t-1} }\cdot \mathbf{x}_t.
\end{aligned}
\end{equation}
Under the framework of Classifier-Free Guidance, when we consider substituting the gradient of data log-likelihood $\nabla_{x_t}\log q(\mathbf{x}_t|c)$ with the CFG score function $s^{cfg}_{\theta,\omega}(\mathbf{x}_t, c)$ to estimate $\mu^*_t$ according to Eq. \ref{optimal_mu}.  The derivation of this estimation lies in the expression:
\begin{equation}
\begin{aligned}
      L(\omega) &\triangleq \left\|s^{cfg}_{\theta,\omega}(\mathbf{x}_t,c) - \nabla_{x_t}\log q(\mathbf{x}_t|c)\right\|^2 \\ 
      &=\left\|\omega \cdot s_\theta(\mathbf{x}_t, c) + (1-\omega) \cdot s_\theta(\mathbf{x}_t, \varnothing) - \nabla_{x_t}\log q(\mathbf{x}_t|c)\right\|^2.
\end{aligned}
\end{equation}    
By treating $\omega$ as a variable, we can obtain the minimum value of $L(\omega)$ as follows:
\begin{equation}
\begin{aligned}
    \omega^* &= \argmin_{\omega} L(\omega)\\
    &=\mathbf{E}_{x_t}\left[\frac{\left(s_{\theta}(\mathbf{x}_t, c) - s_{\theta}(\mathbf{x}_t, \varnothing)\right)\left(\nabla_{x_t}\log q(\mathbf{x}_t|c)-s_{\theta}(\mathbf{x}_t, \varnothing)\right)}{\left(s_{\theta}(\mathbf{x}_t, c) - s_{\theta}(\mathbf{x}_t, \varnothing)\right)^2}\right]\\
    &=\mathbf{E}_{x_t}\left[\frac{\left(s_{\theta}(\mathbf{x}_t, c) - s_{\theta}(\mathbf{x}_t, \varnothing)\right)\left( s_{\theta}(\mathbf{x}_t, c) + e_{t,c}-s_{\theta}(\mathbf{x}_t, \varnothing)\right)}{\left(s_{\theta}(\mathbf{x}_t, c) - s_{\theta}(\mathbf{x}_t, \varnothing)\right)^2}\right]\\
    &= 1 + \mathbf{E}_{x_t}\left[\frac{\left(s_{\theta}(\mathbf{x}_t, c) - s_{\theta}(\mathbf{x}_t, \varnothing)\right)e_{t,c}}{\left(s_{\theta}(\mathbf{x}_t, c) - s_{\theta}(\mathbf{x}_t, \varnothing)\right)^2}\right].
\end{aligned}
\end{equation}
Then we reach the conclusion:
\begin{equation}
    \omega^*= 1 + \mathbf{E}_{x_t}\left[\frac{\left(s_{\theta}(\mathbf{x}_t, c) - s_{\theta}(\mathbf{x}_t, \varnothing)\right)e_{t,c}}{\left(s_{\theta}(\mathbf{x}_t, c) - s_{\theta}(\mathbf{x}_t, \varnothing)\right)^2}\right],
\end{equation}
where $e_{t,c}\triangleq \nabla_{x_t}\log q(\mathbf{x}_t|c) - s_{\theta}(\mathbf{x}_t, c)$ indicates the error between $\nabla_{x_t}\log q(\mathbf{x}_t|c)$ and $s_{\theta}(\mathbf{x}_t, c)$ at time step $t$.

\subsection{Detailed process of Sec. 4.2}
Following the setting of theoretical known diffusion process $q$ and predicted reverse process $p_{\theta}$ \cite{song2020score}, we consider metricing the Kullback-Leibler Divergence (KL Divergence, $D_{KL}$) between these two: 
\begin{equation}
\begin{aligned}
    &D_{KL}[p_{\theta}(\mathbf{x}_{0:T-1}|\mathbf{x}_T)||q(\mathbf{x}_{0:T-1}|\mathbf{x}_T)]\\
    =& \int p_{\theta}(\mathbf{x}_{0:T-1}|\mathbf{x}_T)\log \frac{p_{\theta}(\mathbf{x}_{0:T-1}|\mathbf{x}_T)}{q(\mathbf{x}_{0:T-1}|\mathbf{x}_T)} d \mathbf{x}_{0:T-1}\\
    =& \int p_{\theta}(\mathbf{x}_{0:T-1}|\mathbf{x}_T)\log \frac{q(\mathbf{x}_{0:T-1},\mathbf{x}_T)p_{\theta}(\mathbf{x}_{0:T-1}|\mathbf{x}_T)}{q(\mathbf{x}_{0:T-1}|\mathbf{x}_T)q(\mathbf{x}_{0:T-1},\mathbf{x}_T)} d\mathbf{x}_{0:T-1}\\
    =& \int p_{\theta}(\mathbf{x}_{0:T-1}|\mathbf{x}_T)\log q(\mathbf{x}_T) d \mathbf{x}_{0:T-1} \\
    &- \int p_{\theta}(\mathbf{x}_{0:T-1}|\mathbf{x}_T)\frac{q(\mathbf{x}_{0:T-1},\mathbf{x}_T)}{p_{\theta}(\mathbf{x}_{0:T-1}|x_T)}d \mathbf{x}_{0:T-1}\\
    =&\log q(\mathbf{x}_T) - \int p_{\theta}(\mathbf{x}_{0:T-1}|\mathbf{x}_T)\frac{q(\mathbf{x}_{0:T-1},\mathbf{x}_T)}{p_{\theta}(\mathbf{x}_{0:T-1}|\mathbf{x}_T)}d \mathbf{x}_{0:T-1}.
\end{aligned}
\end{equation}
As a result:
\begin{equation}
    \begin{aligned}
        &\min D_{KL}[p_{\theta}(\mathbf{x}_{0:T-1}|\mathbf{x}_T)||q(\mathbf{x}_{0:T-1}|\mathbf{x}_T)] \\
        = &\max \int p_{\theta}(\mathbf{x}_{0:T-1}|\mathbf{x}_T)\log \frac{q(\mathbf{x}_{0:T-1},\mathbf{x}_T)}{p_{\theta}(\mathbf{x}_{0:T-1}|\mathbf{x}_T)}d \mathbf{x}_{0:T-1}.\\
    \label{opt_targte}
    \end{aligned}
\end{equation}
Following \cite{song2020score}, the predicted reverse process can be expressed as :
\begin{equation}
    p_{\theta}(\mathbf{x}_{0:T-1}|\mathbf{x}_T) = \Pi^T_{t=1} p_{\theta}(\mathbf{x}_{t-1}|\mathbf{x}_t).
\end{equation}
Then Eq.\ref{opt_targte} can be expressed as:
\begin{equation}
\begin{aligned}
    &\max \int p_{\theta}(\mathbf{x}_{0:T-1}|\mathbf{x}_T) \log \frac{q(\mathbf{x}_{0:T-1},\mathbf{x}_T)}{p_{\theta}(\mathbf{x}_{0:T-1}|\mathbf{x}_T)}d \mathbf{x}_{0:T-1}\\
    =& \max ~\scalebox{1.4}[1.4]{[}\int p_{\theta}(\mathbf{x}_{0:T-1}|\mathbf{x}_T)\log q(\mathbf{x}_{0:T-1},\mathbf{x}_T) d \mathbf{x}_{0:T-1} \\
    &~~~~~~~~- \int p_{\theta}(\mathbf{x}_{0:T-1}|\mathbf{x}_T)\log p_{\theta}(\mathbf{x}_{0:T-1}|\mathbf{x}_T)  d \mathbf{x}_{0:T-1}\scalebox{1.4}[1.4]{]}\\
    =& \max ~\scalebox{1.4}[1.4]{[}\int p_{\theta}(\mathbf{x}_{0:T-1}|\mathbf{x}_T)\log q(\mathbf{x}_{0:T-1}|\mathbf{x}_T)q(\mathbf{x}_T) d \mathbf{x}_{0:T-1}\\
    &~~~~~~~~- \int \Pi^T_{t=1} p_{\theta}(\mathbf{x}_{t-1}|\mathbf{x}_t) \log \Pi^T_{t=1} p_{\theta}(\mathbf{x}_{t-1}|\mathbf{x}_t) d\mathbf{x}_{0:T-1}\scalebox{1.4}[1.4]{]}\\
    =&\max ~\scalebox{1.4}[1.4]{[}\log q(\mathbf{x}_T) + \mathbf{E}_{p_{\theta}}\log q(\mathbf{x}_{0:T-1}|\mathbf{x}_T) \\
    &~~~~~~~~- \Sigma _{t=1}^T \mathbf{E}_{p_{\theta}}\log p_{\theta}(\mathbf{x}_{t-1}|\mathbf{x}_t)\scalebox{1.4}[1.4]{]}.
\end{aligned}    
\end{equation}
The above optimization problem can be tackled according to the philosophy of mean–field approximation: the optimization over the entire horizon $t \in [0,T]$ can be decomposed into a collection of single time steps $t \in \{1,..., T\}$ that can be solved iteratively.
For a specific $t \in \{1,..., T\}$, we have:
\begin{equation}
\begin{aligned}
    & \argmax_{p_{\theta}(\mathbf{x}_{t-1}|\mathbf{x}_t)} \scalebox{1.4}[1.4]{[}\log q(\mathbf{x}_T) + \mathbf{E}_{p_{\theta}}\log q(\mathbf{x}_{0:T-1}|\mathbf{x}_T) \\
    &~~~~~~~~~~~~~~~~~- \Sigma _{t=1}^T \mathbf{E}_{p_{\theta}}\log p_{\theta}(\mathbf{x}_{t-1}|\mathbf{x}_t)\scalebox{1.4}[1.4]{]}\\
    =&\argmax_{p_{\theta}(\mathbf{x}_{t-1}|\mathbf{x}_t)} \scalebox{1.4}[1.4]{[}\log q(\mathbf{x}_T) + \mathbf{E}_{p_{\theta}}\log q(\mathbf{x}_{0:T-1/t-1}, \mathbf{x}_{t-1}|\mathbf{x}_T) \\
    &~~~~~~~~~~~~~~~~~- \Sigma _{t=1}^T \mathbf{E}_{p_{\theta}}\log p_{\theta}(\mathbf{x}_{t-1}|\mathbf{x}_t)\scalebox{1.4}[1.4]{]}\\
    =&\argmax_{p_{\theta}(\mathbf{x}_{t-1}|\mathbf{x}_t)} \scalebox{1.4}[1.4]{[}\log q(\mathbf{x}_T) + \mathbf{E}_{p_{\theta}}\log q(\mathbf{x}_{t-1}|\mathbf{x}_{0:T-1/t-1}, \mathbf{x}_T) \\
    &~~~~~~~~~~~~~~~~~+\mathbf{E}_{p_{\theta}}\log q(\mathbf{x}_{0:T-1/t-1}|\mathbf{x}_T)\\
    &~~~~~~~~~~~~~~~~~- \Sigma _{t=1}^T \mathbf{E}_{p_{\theta}}\log p_{\theta}(\mathbf{x}_{t-1}|\mathbf{x}_t)\scalebox{1.4}[1.4]{]}\\
    =&\argmax_{p_{\theta}(\mathbf{x}_{t-1}|\mathbf{x}_t)} \scalebox{1.4}[1.4]{[}\mathbf{E}_{p_{\theta}}\log q(\mathbf{x}_{t-1}|\mathbf{x}_{0:T-1/t-1}, \mathbf{x}_T) \\
    &~~~~~~~~~~~~~~~~~-  \mathbf{E}_{p_{\theta}}\log p_{\theta}(\mathbf{\mathbf{x}}_{t-1}|\mathbf{x}_t)\scalebox{1.4}[1.4]{]}\\
    =& \argmax_{p_{\theta}(\mathbf{x}_{t-1}|\mathbf{x}_t)} \scalebox{1.4}[1.4]{[}\int p_{\theta}(\mathbf{x}_{t-1}|x_t)\log q(\mathbf{x}_{t-1}|\mathbf{x}_{0:T-1/t-1}, \mathbf{x}_T) d\mathbf{x}_{t-1} \\
    &~~~~~~~~~~~~~~~~~- \int p_{\theta}(\mathbf{x}_{t-1}|\mathbf{x}_t)\log p_{\theta}(\mathbf{x}_{t-1}|\mathbf{x}_t)d\mathbf{x}_{t-1}\scalebox{1.4}[1.4]{]}.
\end{aligned}
\end{equation}
Then we reach the conclusion with the Lagrange Multiplier Method:
\begin{equation}
    p_{\theta^*}(\mathbf{x}_{t-1}|x_t) = \mathbf{E}_{p_{\theta^*}} q(\mathbf{x}_{t-1}|\mathbf{x}_{0:T-1/t-1}, \mathbf{x}_T),
\end{equation}
where the left side indicates the predicted distribution from the model given $\mathbf{x}_t$ and the right side means the theoretical distribution of the diffusion process given the same $\mathbf{x}_t$. 
Considering the setting from \cite{song2020score}, the convergence target of every step should be:
\begin{equation}
    L = \left\|\epsilon^{cfg}_{\theta,\omega}(\mathbf{x}_t,c) - \epsilon\right\|^2,
\end{equation}
here $\epsilon^{cfg}_{\theta,\omega}(\mathbf{x}_t,c)$ means the weight score function with CFG and $\epsilon\sim \mathcal{N}(0, \mathbf{I})$. In this work, we apply the simple gradient descent method to accomplish this optimization task, with a step size of $5e-2$ and a threshold of $1e-3$. The proposed algorithm is expressed as:
\begin{algorithm}[h]
\caption{\model: Optimizing Diffusion Models with Iterative Error Reduction}
\label{alg:optimization1}
\begin{algorithmic}[1]
\STATE Input: classifier-free guidance weight: $\omega$ 
\STATE ~~~~~~~~~~~gradient scale: $\eta = 5e-2$
\STATE ~~~~~~~~~~~convergence threshold: $1e-3$
\STATE $\mathbf{x}_T \sim \mathcal{N}(0, \mathbf{I})$
\FOR{$t = T$ to $1$}
    \STATE $ \epsilon^{\text{cfg}}_{\theta, \omega} (\mathbf{x}_t, c)= \epsilon_{\theta}(\mathbf{x}_t, \varnothing) + \omega \cdot \left(\epsilon_{\theta}(\mathbf{x}_t, c) - \epsilon_{\theta}(\mathbf{x}_t, \varnothing)\right)$
    \WHILE{not convergence}
        \STATE $\epsilon \sim \mathcal{N}(0, \mathbf{I})$
        \STATE L = $\left \| \epsilon^{cfg}_{\theta, \omega}(\mathbf{x}_t, c)-\epsilon \right  \|^2$
        \STATE $\epsilon^{cfg}_{\theta, \omega}(\mathbf{x}_t, c) = \epsilon^{cfg}_{\theta, \omega}(\mathbf{x}_t, c)+\eta \cdot \nabla_{\epsilon^{cfg}_{\theta, \omega}(\mathbf{x}_t, c)} $L
    \ENDWHILE
    \STATE $x_{t-1}$ = DDIM Sampler$\left(\mathbf{x}_t, \epsilon^{cfg}_{\theta, \omega}(\mathbf{x}_t, t)\right)$
\ENDFOR
\STATE \textbf{return} $\mathbf{x}_0$
\end{algorithmic}
\end{algorithm}

\section{Additional quantitative resuts}
We also assess our approach for \textit{unconditional sampling} with the Latent Diffusion Model (LDM) trained on datasets including CelebA~\cite{liu2015faceattributes}, LSUN-Church, and LSUN-Bedroom~\cite{yu2015lsun}. We extend our analysis using the Predictor-Corrector sampling strategy from~\cite{song2020score}, comparing results with the official pre-trained model. 

\begin{table}[h]
\caption{Numerical results of unconditional sampling on CelebA (C-128, number for various image sizes)dataset, Lsun-bedroom (L-b), and Lsun-church (L-c) datasets versus latent diffusion method (LDM). We generate 50K samples for each comparison.}
\centering
\begin{tabular}{cccccc}
\toprule
Method &Val. set & IS $\uparrow$ &KID $\downarrow$ & Prec. $\uparrow$ & Rec. $\uparrow$ \\
\toprule
 LDM&C-128&  3.27 & 0.0661 & 0.052 & 0.023 \\
 \model &C-128&  \textbf{3.28}& \textbf{0.0658} & \textbf{0.052} & \textbf{0.024} \\
 \hline
 LDM &C-256&  3.27& 0.0184 & 0.468 & 0.316 \\
 \model &C-256&  \textbf{3.28}& \textbf{0.0183} & \textbf{0.470} & \textbf{0.320} \\
  \hline
 LDM &C-512&  3.27& 0.0060 & 0.725 & 0.519 \\
 \model &C-512&  \textbf{3.28}& \textbf{0.0058} & \textbf{0.727} & \textbf{0.526} \\
  \hline
 LDM &C-1024& 3.27& 0.0075 & 0.729 & 0.515 \\
 \model &C-1024& \textbf{3.28}& \textbf{0.0073} & \textbf{0.734} & \textbf{0.523} \\
 \hline
  LDM& L-b &  2.23 & \textbf{0.0032} & 0.724 & 0.458 \\
 \model & L-b  &  \textbf{2.23}& 0.0035 & \textbf{0.724} & \textbf{0.463} \\
 \hline
 LDM & L-c &  2.70& 0.0062 & 0.766 & 0.453 \\
 \model & L-c  &  \textbf{2.72}& \textbf{0.0060} & \textbf{0.774} & \textbf{0.453} \\
\bottomrule
\end{tabular}
\label{tab:latent-CelebA}
\end{table}

\paragraph{Unconditional Image Generation.} We also present a quantitative comparison for unconditional sampling using LDM on datasets including CelebA, LSUN-Church, and LSUN-Bedroom. Due to the varying image sizes in CelebA (e.g., $128^2$, $256^2$, $512^2$, and $1024^2$), we use 'CelebA-128' to denote the corresponding generated image size. Since image size can impact metric outcomes, we conduct comparisons across all sizes for a comprehensive perspective. As shown in Table~\ref{tab:latent-CelebA}, our approach outperforms LDM across multiple perceptual metrics, including IS, KID, Precision, and Recall. For instance, on CelebA-256, our \model achieves an IS score of 3.28, surpassing the baseline. Additionally, it achieves Precision and Recall scores of 0.470 and 0.320, respectively. Similarly, on the LSUN-Church validation dataset, our method achieves an IS score of 2.72, with Precision and Recall scores of 0.774 and 0.453, outperforming the baseline. These results demonstrate that our method achieves superior performance in unconditional sampling tasks.




\paragraph{Discussion on the computation cost of this work.}
As a training-free optimization method, it is necessary to discuss the computation cost of this work. First of all, since this method does not involve a training procedure, it has hardware resource requirements for model training. As illustrated in the scripts, this work aims to reduce the ``training-inference gap'' of diffusion models with a gradient-based method. This implies that for a model, when this gap is very small, the computational cost of optimization will also be extremely low. As tabulated in Tab.~\ref{tab:latent-CelebA}, despite being an unconditional generation instance, the model demonstrates negligible intrinsic bias as evidenced by the high fidelity of its generated outputs. Consequently, the magnitude of improvement remains modest, while the optimization procedure incurs minimal computational latency. On the other hand, for conditional generation tasks (such as text-to-image generation), when the model's predicted values under a fixed input condition exhibit significant error from theoretical values, the iterative optimization strategy of this method can reduce such error, thereby enhancing generation quality. Leveraging gradient-based characteristics, the proposed method achieves high computational efficiency. In this scenario, the computational cost is deemed acceptable considering the improved output quality.

\section{More visible relusts}
\subsection{Visible results on Text-to-Image Generation task}
\begin{figure*}[htb]
\begin{center}
   \includegraphics[width=0.9\linewidth]{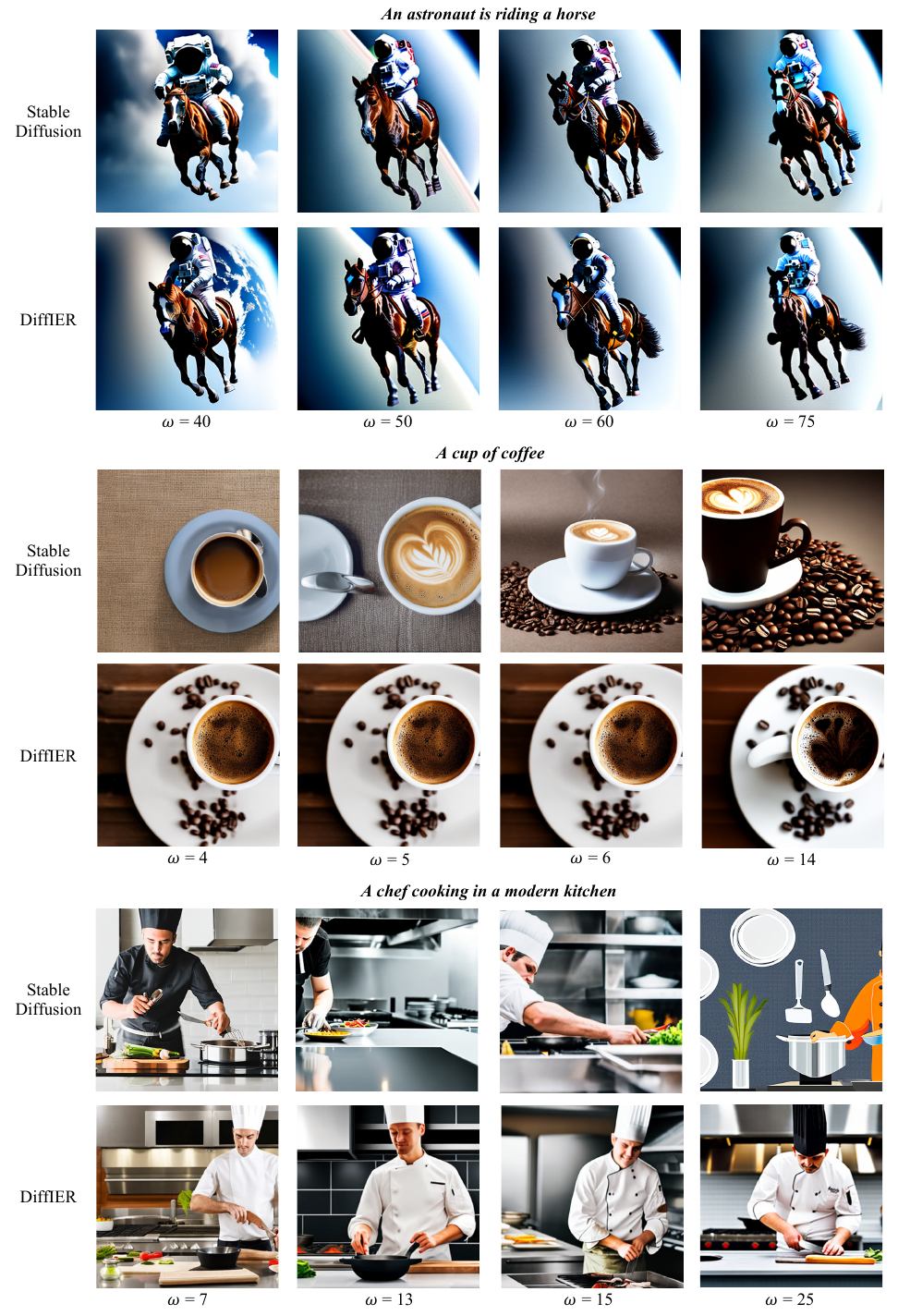}
\end{center}
   \caption{Results on Text-to-Image Generation task, compared with Stable Diffusion. }
\label{fig:supply_stable_diffusion}
\end{figure*}

We illustrate more visible results on the Text-to-Image Generation task in Fig.~\ref{fig:supply_stable_diffusion}.

\subsection{Visible results on Image Super-Resolution task}
\begin{figure*}[htb]
\begin{center}
   \includegraphics[width=0.86\linewidth]{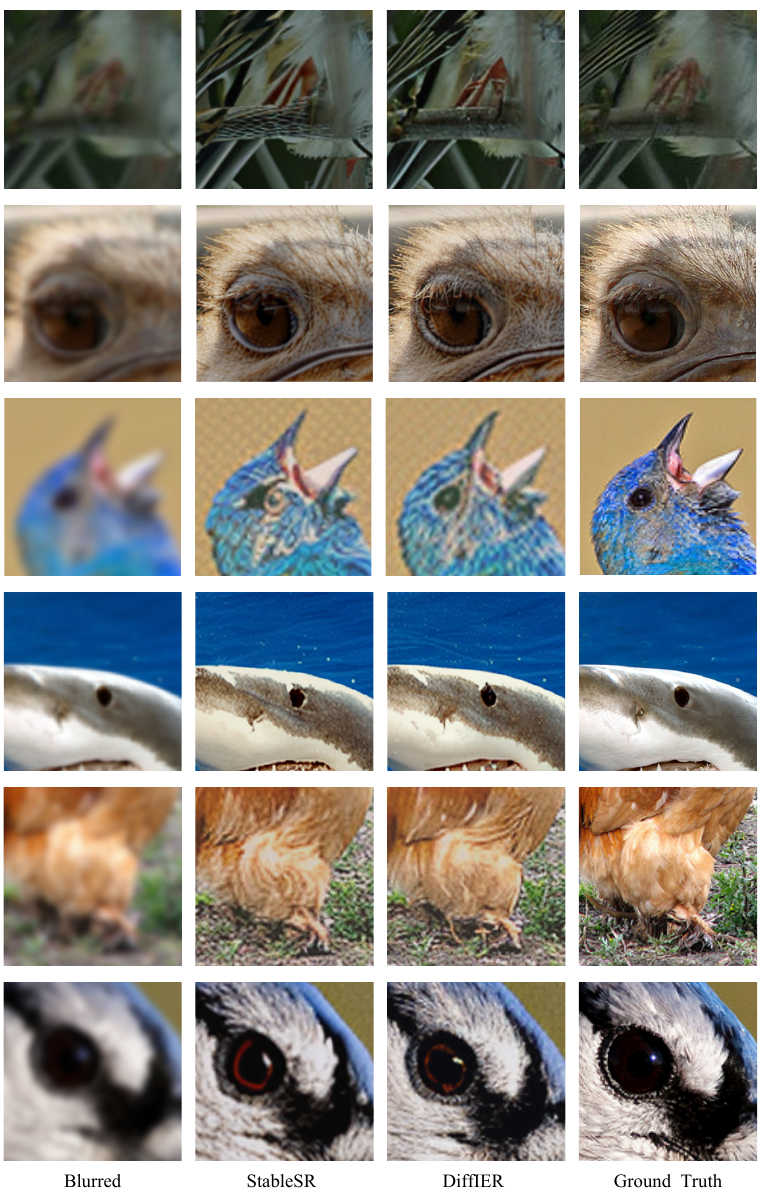}
\end{center}
   \caption{Results on Image Super-Resolution task, compared with StableSR.}
\label{supplyfig:supple_sr}
\end{figure*}

We illustrate more visible results on the Image Super-Resolution task in Fig.~\ref{supplyfig:supple_sr}.

\subsection{Visible results on Text-to-Speech Generation task}
\begin{figure*}[htb]
\begin{center}
   \includegraphics[width=0.9\linewidth]{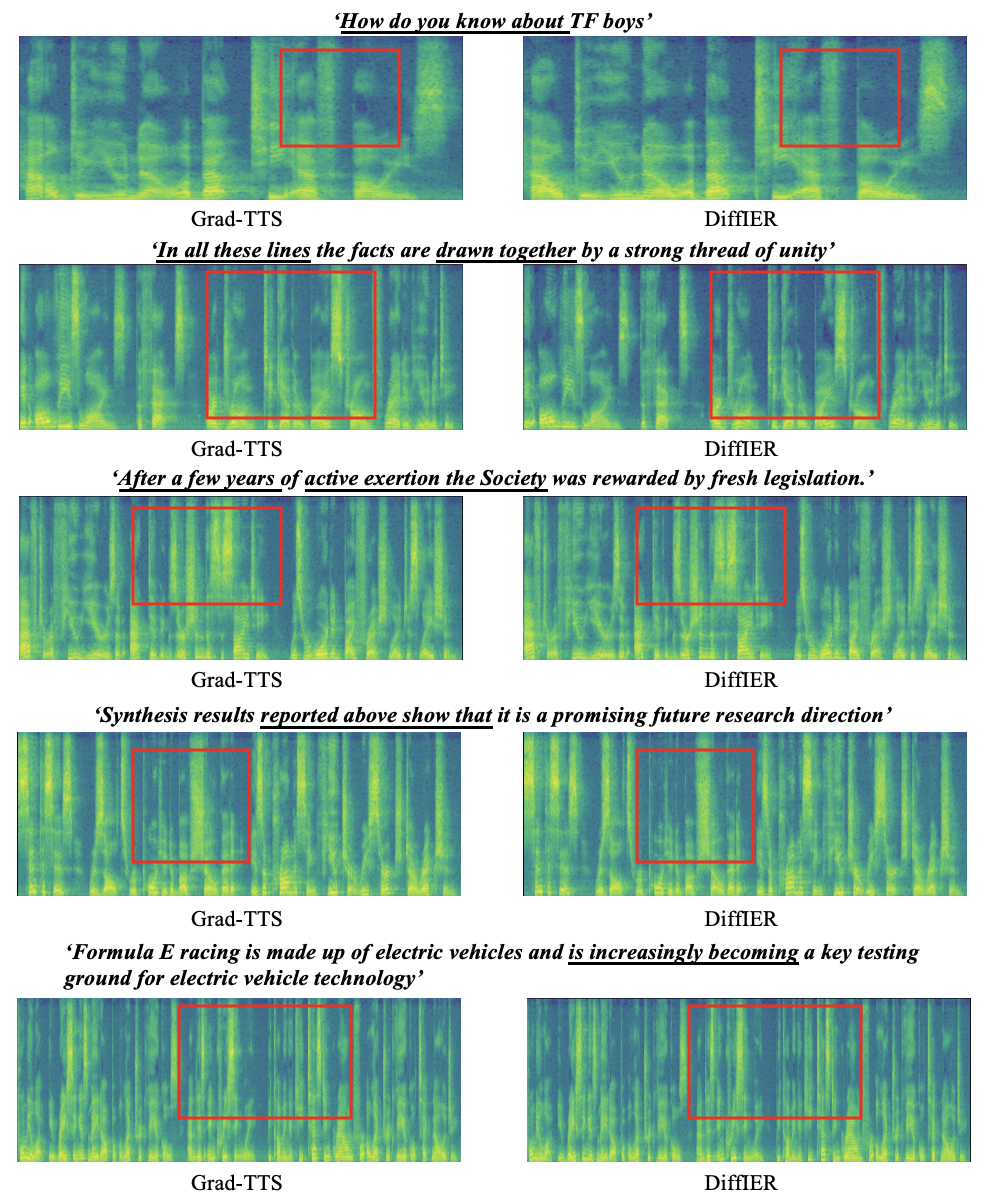}
\end{center}
   \caption{Results on Text-to-Speech Generation task, compared with Grad-TTS. The underlined segments of the text demonstrate significant differences in linguistic and stylistic characteristics.}
\label{fig:supple_speech}
\end{figure*}

We illustrate more visible results on the Text-to-Speech Generation task in Fig.~\ref{fig:supple_speech}.

\end{document}